\documentclass[10pt,twocolumn,letterpaper]{article}

\usepackage{cvpr}
\usepackage{times}
\usepackage{epsfig}
\usepackage{graphicx}
\usepackage{amsmath}
\usepackage{amssymb}
\usepackage{appendix}
\usepackage[breaklinks=true,bookmarks=false,hidelinks]{hyperref}% breaklinks=true,bookmarks=false

\cvprfinalcopy % *** Uncomment this line for the final submission

\begin{document}

%%%%%%%%% TITLE
\title{Seeing in the dark with recurrent convolutional neural networks}

\author{Till S. Hartmann\\
Harvard Medical School\\
{\tt\small till@hms.harvard.edu}\\
\\
\today
}

\maketitle

%%%%%%%%% ABSTRACT
\begin{abstract}
Classical convolutional neural networks (cCNNs) are very good at categorizing objects in images. But, unlike human vision which is relatively robust to noise in images, the performance of cCNNs declines quickly as image quality worsens. Here we propose to use recurrent connections within the convolutional layers to make networks robust against pixel noise such as could arise from imaging at low light levels, and thereby significantly increase their performance when tested with simulated noisy video sequences. We show that cCNNs classify images with high signal to noise ratios (SNRs) well, but are easily outperformed when tested with low SNR images (high noise levels) by convolutional neural networks that have recurrency added to convolutional layers, henceforth referred to as gruCNNs. Addition of Bayes-optimal temporal integration to allow the cCNN to integrate multiple image frames still does not match gruCNN performance. Additionally, we show that at low SNRs, the probabilities predicted by the gruCNN (after calibration) have higher confidence than those predicted by the cCNN. We propose to consider recurrent connections in the early stages of neural networks as a solution to computer vision under imperfect lighting conditions and noisy environments; challenges faced during real-time video streams of autonomous driving at night, during rain or snow, and other non-ideal situations.
\end{abstract}

%%%%%%%%% BODY TEXT
\section{Introduction}
	\begin{table*}[!]
	\centering
	\begin{tabular}{ |c|c|c| } 
		
		\hline
		\textbf{\#} & \textbf{Layer type} & \textbf{Parameters} \\ \hline \hline
		1 & Conv layer + ReLU & Kernel size: 3$\times$3, number channels: 96 (no recurrency) or 32 (recurrency)  \\ \hline
		2 & Conv layer + ReLU & Kernel size: 3$\times$3, number channels: 96 (no recurrency) or 32 (recurrency)  \\ \hline
		3 & Max pooling & 2$\times$2 \\ \hline
		4 & Dropout & 25\% \\ \hline
		5 & Batch normalization & Time distributed -- for each frame individually  \\ \hline
		6 & Conv layer + ReLU & Kernel size: 3$\times$3, number channels: 192 (no recurrency) or 64 (recurrency)  \\ \hline
		7 & Conv layer + ReLU & Kernel size: 3$\times$3, number units: 192 (no recurrency) or 64 (recurrency)  \\ \hline
		8 & Max pooling & 2$\times$2 \\ \hline
		9 & Dropout & 25\% \\ \hline
		10 & Flatten & Time distributed -- for each frame individually \\ \hline
		11 & Fully connected + ReLU & Number channels: 1536 (no recurrency) or 512 (recurrency) \\ \hline
		12 & Dropout & 50\% \\ \hline
		13 & Fully connected + softmax & 10 output channels (independent of recurrency) \\ \hline
	\end{tabular}
	\caption{Default model architecture tested.
	\label{tab:model}}
\end{table*}
Classical convolutional neural networks (cCNNs) classify objects from static, high quality images exceptionally well. However, in many applications (autonomous driving, robotics), imaging data is rarely derived from photographs taken under excellent lighting conditions, but rather from images with occluded objects, motion distortion, low signal to noise ratios—resulting from poor image quality or low light levels—and frequently streaming over time (video). cCNNs have been used successfully for image denoising to enhance the quality of single photographs \cite{Zhang2017,Chen2018}; outperforming traditional computer vision approaches such as block-matching and 3-d filtering (BM3D) \cite{Dabov2007}. However, these networks rely on standard feedforward architectures and cannot combine information from multiple frames of a video sequence. As such, cCNNs are easily outperformed by humans if even small levels of noise are added to images \cite{Geirhos2017}. It is clear that systems that rely on computer vision to navigate the real world should be able to do so in all kind of conditions, from bright mid-day sun, rainy days, or at night.

Generally, neuronal latencies increase in visual cortex in response to stimuli with low contrast \cite{Maunsell1999}. In a study of inferior temporal cortex, researchers found that the increased response time for low contrast stimuli was dependent on the selectivity of individual neurons---ranging from tens of milliseconds for broadly tuned neurons, to 100--200 ms for very selective neurons \cite{Oram2010}. This suggests that despite their broadly-tuned neighbors being active earlier, these selective neurons do not become active until additional information has been integrated by previous stages of the visual system. Furthermore, humans without a viewing duration constraint have trouble identifying occluded images; a task a cCNN such as AlexNet \cite{Krizhevsky2012} does not perform particularly well, either \cite{Tang2018}. Allowing the human observers more time, or adding recurrent connections to the last layer (fc6) of AlexNet improves performance for both \cite{Tang2018}, indicating how beneficial recurrent connections can be. We hypothesize that when responding to ambiguous stimuli, such as in low contrast conditions or occlusion, the neural circuit permits more time for recurrent processing before selective neurons---presumably important for decision-making---become active.

The classical feedforward CNNs were inspired by the primate visual system \cite{Fukushima1982}. Yet, this architecture ignores recurrent connections, which are abundant in the primate visual system. We believe that the visual system---especially at low contrast/low SNR values---acts as both a CNN and a recurrent neural network, in which local recurrent connections can integrate information at every layer. To our knowledge, no one has used recurrent convolutional layers to boost classification performance or other computer vision tasks at low SNRs. Others have added recurrent connections in early convolutional layers of CNNs \cite{Liang2015,Finn2016}, which improved performance in ImageNet classification \cite{Nayebi2018} and aided in digit clutter analysis \cite{Spoerer2017}. Here, we show that cCNN performance using a standard cCNN architecture breaks down in the low-SNR regime, while a recurrent CNN can still classify relatively well. We systematically compare the integration performance of recurrent CNNs (gruCNNs) to a Bayes-optimal temporal integration of cCNNs, and show that integration at early processing stages with recurrent connections is significantly improved.
\begin{figure*}
	\begin{center}
		\includegraphics{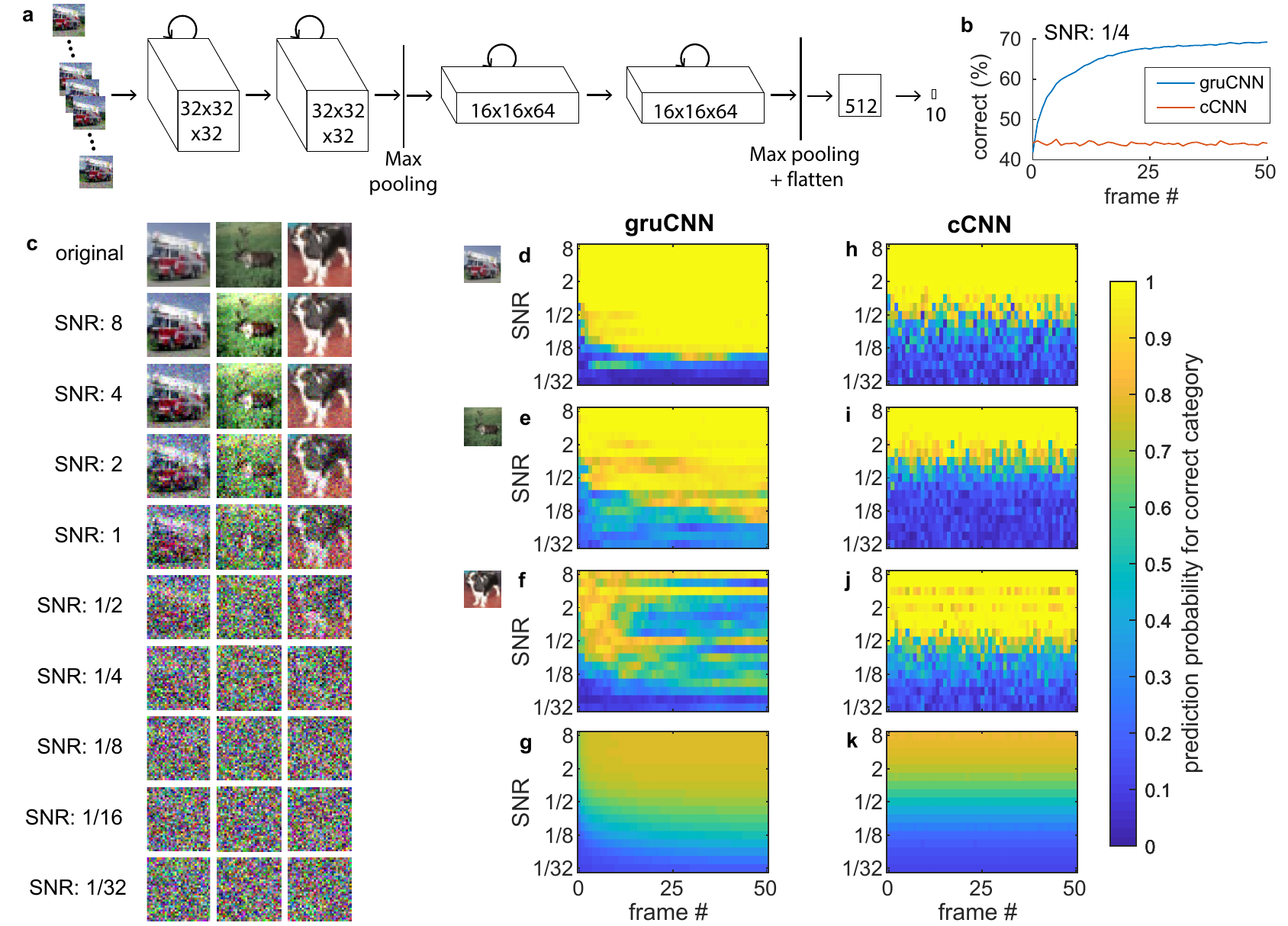}
	\end{center}
	\caption{\textbf{Architecture and example data.} a) Architecture of gruCNN. Each channel’s activity depends on both the current input as well as the previous state. b) Classification performance of example gruCNN and cCNN when all test sequences had an SNR of $1/4$. c) Original image and image with different SNRs for a firetruck (category truck) a reindeer (category deer), and a dog, shown without jitter. d--k) Color coded predicted probabilities (output of \textit{softmax}) of the correct (positive) image category for gruCNN (d--g) and cCNN (h--k). Horizontal axes show predicted probabilities over 51 frames, vertical axes over a range of SNRs. d) \& h) and e) \& i) correspond to performance in the fire truck and reindeer examples, respectively. The predictive probability at low SNRs continue improving over frames for the gruCNN predictions, but are relatively constant for the cCNN. f) \& j) Data for the third example (the dog), in which the gruCNN fails (which is rare) while the cCNN predicts the category correctly at most SNRs. The average predicted probability for correct (positive) image category for all 10,000 test images is displayed in g) \& k). }
	\label{fig:example}
\end{figure*}

\section{Training and model architecture}
To simulate video sequences filmed in different light conditions we augmented the CIFAR-10 dataset \cite{Krizhevsky2009} by adding noise and jitter to the original image over time. 

\begin{figure*}
	\begin{center}
		\includegraphics{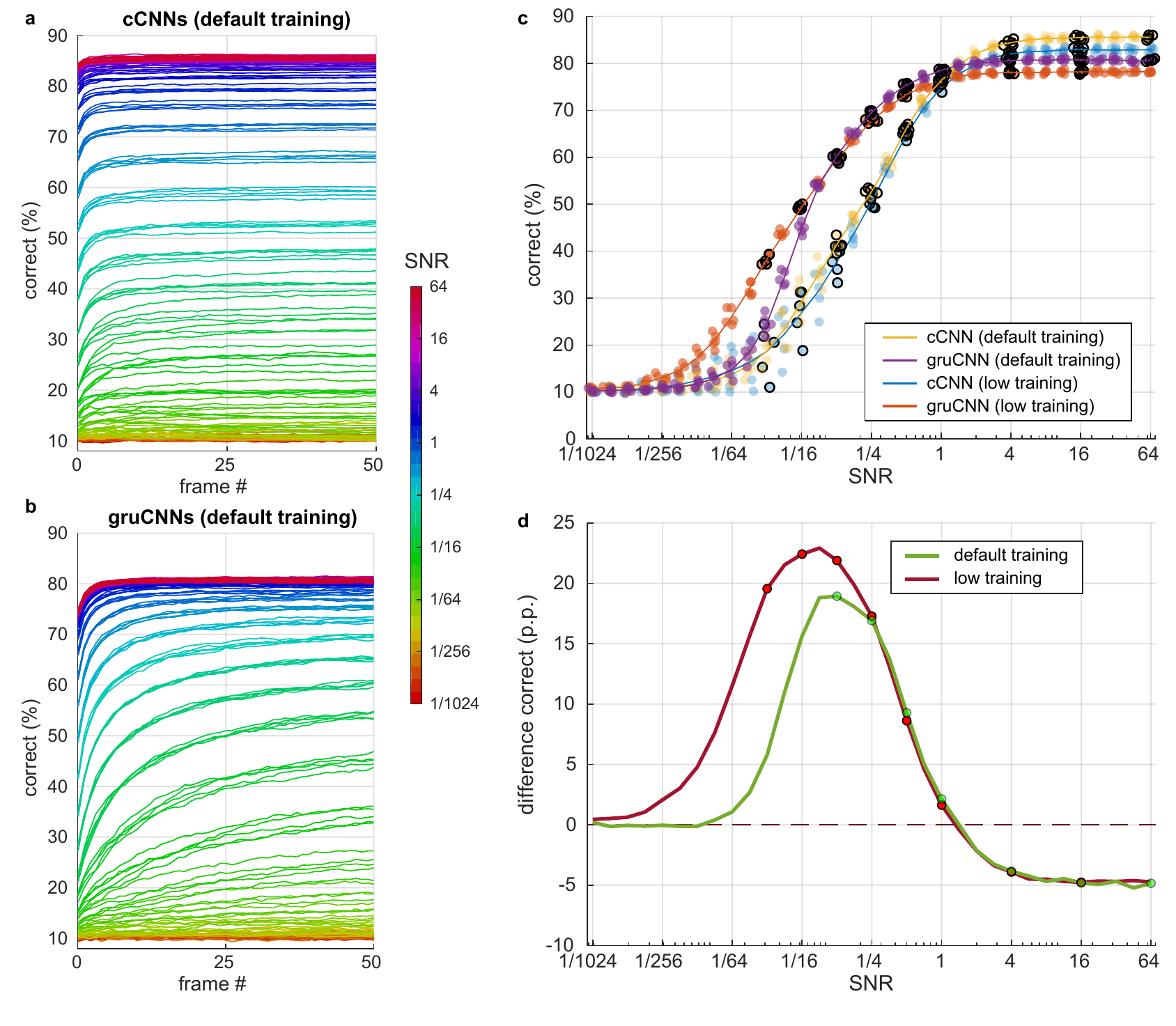}
	\end{center}
	\caption{\textbf{Detailed comparison of cCNN with Bayesian inference and gruCNN performance over a large range of SNR levels.} Each model architecture was tested after training with slightly higher SNRs (default training) and after training with slightly lower SNRs (low training). a) \& b) Percent correct over the course of 51 frames for different SNRs (color coded) using default training for a) the cCNN (with Bayesian Inference) and b) the gruCNN. c) Dots: correct classification for the model architectures at the last frame. Jitter in SNR values was added to increase readability of plots, but was not in the data. Lines: mean performance of the five models per architecture. d) Mean performance of gruCNNs minus mean performance of cCNNs for models trained with default and lower SNRs (green and red, respectively). SNR levels used during training are indicated by dots. }
	\label{fig:manySNR}
\end{figure*}

\subsection{Image sequence creation}
Before creating image sequences with different SNRs, we preprocessed the 60,000 images in the CIFAR-10 dataset by subtracting the global mean from all images and normalizing them by the global standard deviation of all image pixels. 
For training, we used the same image 26 times, stacking it into an image sequence. We randomly jittered each picture’s position within the stack by up to three pixels in each direction to induce spatial uncertainty over time. Missing pixels were filled with their nearest neighbor’s values. Each training image sequence was generated such that all images within one sequence had a randomly chosen SNR of 64, 16, 4, 1, 1/2, 1/4, or 1/8 (default training). Each image sequence was normalized after adding the noise. The image sequences were designed to mimic different environmental conditions, from \eg bright daylight (high SNR) to a foggy night (low SNR). 

\subsection{Combination of recurrency with convolutions: The gruCNN layer}
We were inspired by the combination of the well-known long-short term memory units (LSTM \cite{Hochreiter1997,Gers2000}) with convolutional layers by Shi \etal\cite{Shi2015}. We derived the gruCNN layer by modifying Shi’s implementation using the gated recurrent units (GRU) \cite{Cho2014} instead of the LSTM.
GRUs have fewer parameters than LSTMs and no memory cells for long term memory storage, but with otherwise similar properties. 
The hidden state $h_t$ in a gruCNN layer is a 3D tensor, whose last two dimensions are spatial dimensions. In contrast, standard non-convolutional recurrent networks like the LSTM or GRU use full connections with no spatial encoding. The key equations for the gruCNN layers are:
\begin{equation}\label{eq:gru1}
z_t = \sigma \big(W_{zh}*h_{t-1} + W_{zx}*x_t  \big) 
\end{equation}
\begin{equation}\label{eq:gru2}
r_t = \sigma \big(W_{rh}*h_{t-1} + W_{rx}*x_t  \big)
\end{equation}
\begin{equation}\label{eq:gru3}
\tilde{h_t} = \tanh \Big(  W_{hh}* \big(r_t \circ h_{t-1}\big) + W_{hx}*x_t  \Big)
\end{equation}
\begin{equation}\label{eq:gru4}
h_t =  \Big( z_t \circ h_{t-1} + \big( 1-z_t \big) \circ \tilde{h_t}  \Big)
\end{equation}
The hidden state $h_t$ receives gated updates depending on the previous hidden state $h_{t-1}$ and input $x_t$. Hidden state and input are convolved with $3\times3$ sized kernels $W$ and combined as shown in equations \eqref{eq:gru1} through \eqref{eq:gru4}, where $\circ$ is the Hadamard (element-wise) product and $*$ indicates convolutions.
\subsection{Model architecture}
All neural networks described in this manuscript consisted of four convolutional layers (with or without recurrency) and two fully connected layers (see Figure \ref{fig:example}a for the recurrent version, Table  \ref{tab:model} for details). The output of the last fully connected layer is passed through a \textit{softmax} to predict the likelihood of a given  image belonging to one of CIFAR-10’s image categories. We modified the model architecture by adding recurrent connections at the convolutional layers (layers 1, 2 and 6, 7).

\subsection{Neural network training}
Training and testing were performed with the python package keras \cite{Chollet2015} either on desktop computers or on the O2 High Performance Compute Cluster, supported by the Research Computing Group, at Harvard Medical School.
All models were trained with stochastic gradient descent using keras’ implementation of rmsprop \cite{Hinton2010} with an initial learning rate of $10^{-3}$ and a learning rate decay over each update of $10^{-6}$. Models were trained for 100 epochs, \ie each of the 50,000 training images was used 100 times to create an image sequence. A training batch consisted of 64 image sequences of 26 time frames of $32\times32\times3$ image pixels. To train all networks similarly, we used identical training parameters for all models irrespective of recurrent connections. We minimized the cross entropy loss between the correct category and an image frame averaged for all 26 frames, meaning models needed to predict each frame correctly, not just the last. 
We tested the models’ predictions with image sequences derived from 10,000 withheld test images for 51 frames repeatedly with different SNRs. To limit dependence on parameter initialization, each model architecture was trained five times.

\section{Performance of gruCNNs and cCNNs}
We find that recurrent connections in convolutional layers are advantageous in classifying CIFAR-10 image sequences when high levels of noise are present (Figure \ref{fig:example}b, Figure \ref{fig:manySNR}). 

\subsection{Output of gruCNNs and cCNNs}Figure \ref{fig:example}c shows three examples of how image quality deteriorates as we decrease the SNR.  For each of the ten image categories, the models predict the probability that a test image belongs in it.
Figure \ref{fig:example}d--k show the probabilities each model assigned to the correct image category for three example images, and average for all tested images. As seen in the examples of Figure \ref{fig:example}c,d the noisier the image, the longer it takes for gruCNNs to predict a high probability for the correct category, while cCNNs predictions are temporally independent (Figure \ref{fig:example}h,i). In rare instances, the gruCNNs failed (Figure \ref{fig:example}f), while cCNNs performed adequately (Figure \ref{fig:example}j). The average predicted probability for the correct image category for all test images reveals that cCNNs performed better at high SNRs, but at low SNRs gruCNNs classified images more accurately (Figure \ref{fig:example}g,k). Next, we  looked at the models’ correct classification performance \ie percent true positives, a positive being the highest probability predicted for a frame. 

\subsection{Bayesian inference improves cCNN classification performance}
The cCNNs treat every frame as independent, despite every frame being derived from the same original image, with independent noise and jitter added. Bayes inference provides a method to integrate independent observations. We apply it here to try to compensate for the disadvantage cCNNs have compared to networks with recurrent connections. We assume that every updated probability is proportional to the product of the previous frame’s probability (prior) and likelihoods derived from the current frame---the output of the last fully connected layer after \textit{softmax}. This Bayesian inference classification rate is identical to the cCNN prediction for the first frame (the first frame has a flat prior), but improves with additional frame presentations, from 82\% to 85\% correct for SNRs 4 and above (Figure \ref{fig:manySNR}a). Bayesian inference produces better classification performance for all SNRs than the cCNNs predictions without inference. % maybe a conclusion? Bayesian inference bla bla bla

\subsection{Comparing gruCNNs and cCNNs}
To compare performance between the two network architectures, we tested classification error over time. The cCNNs outperform gruCNNs at high SNRs (Figure \ref{fig:manySNR}a,b). However, we noted that the classification error of gruCNNs improves significantly over time, especially for low SNRs (Figure \ref{fig:manySNR}b). This ability of the gruCNNs to integrate additional information, leads to significantly improved performance compared to the cCNNs at low SNRs, despite cCNNs improvement from Bayesian inference. This is clearly visualized when we look at classification performance at frame 50 for all SNR levels (Figure \ref{fig:manySNR}c, yellow line (cCNNs) versus purple line (gruCNNs)). This suggests that for a network to compensate for high noise levels, temporal integration at early processing stages is highly beneficial.

We do not think it is problematic that the gruCNNs underperform at high SNRs as we can choose the appropriate model for any ambient noise level. In order to roughly match the numbers of parameters between the two models, currently all gruCNNs have a third of the input channels per convolutional layer compared to the cCNNs. Thus, we expect that gruCNNs with the same number of input channels per layer to perform equally well at high SNRs. The dataset we chose (CIFAR-10) is too small to test this hypothesis as gruCNNs with increased channel count overfit the training data, however more training images should solve that problem. These data reveal significantly higher classification performance at low SNR levels of gruCNNs compared to the cCNNs.

\subsection{gruCNNs improve performance at low SNRs}
Both network architectures (gruCNN and cCNN) were trained five times with a set of SNRs, typically: 64, 16, 4, 1, 1/2, 1/4, 1/8 (default training). We then tested the effect on classification performance when trained with a lower set of SNRs (16, 4, 1, 1/2, 1/4, 1/8, 1/16, 1/32, low training). Performance at higher SNRs is best for cCNNs with default training and slightly decreased for  cCNNs with low training (Figure \ref{fig:manySNR}c yellow and blue line, respectively). The performance at low SNR values is similar, irrespective of training set; cCNNs did not benefit from the noisier training. As stated above, gruCNNs' performance at high SNRs did not reach levels of cCNNs with default training, and this was further decreased when the low SNR set was used to train (Figure \ref{fig:manySNR}c purple vs red lines). Notably, gruCNNs with low training predicted % did I change time correctly?
even more accurately at low SNRs compared to the gruCNNs with default training, revealing the models’ ability to benefit from noisier training sequences. Figure \ref{fig:manySNR}d compares the mean difference of gruCNNs and cCNNs performance (gruCNN correct \% $-$ cCNN correct \%) for the two SNR training sets. Performance in low SNR regime shows a dramatic improvement from low training, despite already significantly outperforming cCNNs with default training. This shows that the recurrent convolutional layers can learn to classify objects correctly at very low signal levels.

\section{Integration time}

\begin{figure}
	\begin{center}
		\includegraphics{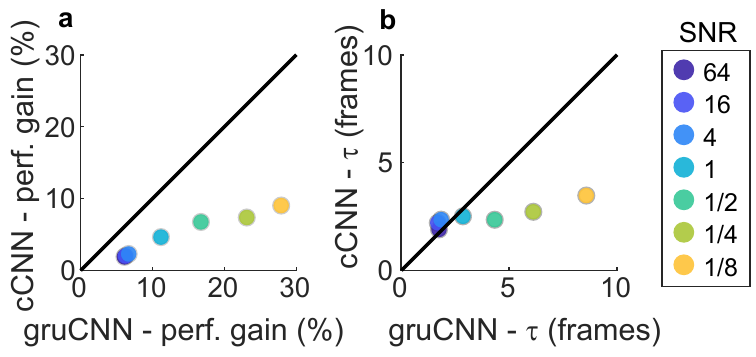}
	\end{center}
	\caption{\textbf{Integration time analysis.} a) Comparison of performance increase between initial ($t=0$) and extrapolated ($t=\infty$) performance for gruCNNs and cCNNs. b) Time constant for performance increase. Colors indicate the tested SNR values.}
	\label{fig:time}
\end{figure}

Both the cCNNs (with Bayesian inference) and  gruCNNs classify images more accurately over time, but the gruCNNs gain more from longer integration and, at low SNR levels, integrate over more frames. We quantified this by fitting the exponential function:
\begin{equation}\label{eq:time}
f(x) = (c-a)  \, e^{-\frac{t}{\tau}}+c
\end{equation}
 to the mean performance over frames (see Figure \ref{fig:manySNR}a, b) for the cCNNs and the gruCNNs. $a$ is the performance increase between the initial performance at $t=0$ and performance at $t=\infty$ (see Figure \ref{fig:time}a), and $\tau$ the time constant (see Figure \ref{fig:time}b). As seen in Figure \ref{fig:time}a, 
 all values lie below the unity line, indicating that for all tested SNR levels gruCNNs’ performance increases more over time than the cCNNs. The gruCNNs' $\tau$ grows consistently larger as SNR decrease, while the cCNNs' $\tau$ only changes marginally. This demonstrates that gruCNNs are integrating additional information for longer periods than cCNNs at low SNRs, a clear advantage when every single frame has low information content due to high noise.

\section{False rejection rate}

Thus far, we have examined the models’ classification performance. Conversely, we can also ask whether image classes rejected by the model were truly absent. To do so, we use the lowest quintile (bottom 20\%) of predicted probabilities to gauge the models ability to rule out categories. This allows us to estimate the false negative rate, influencing how much we can empirically trust each model when it predicts very low probabilities for a given category. In other words, can we believe the model that, when it predicts that there is no truck, that there is actually no truck? Five gruCNNs and five cCNNs were tested with two SNRs, 4 and 1/4. Each image sequence of 51 frames was tested 20 times with different random noise added to the image sequences. Ten thousand test images, ten CIFAR-10 categories, five trained models per architecture and the 20 repetitions resulted in ten million predicted probabilities for each frame per model architecture and SNR level. For each model architecture and frame, we took the lowest quintile of predicted probabilities and labeled these as “rejected”. To estimate the false rejection rate, we counted how many time the correct image category was present within this lowest quintile. A perfect model’s predictions would produce a false rejection rate of 0\%; mere performance at chance would produce a rate of 10\%. Using other percentiles (\eg 1--15\%) to calculate the false rejection rate resulted in qualitatively similar results. However, we chose to examine the false rejection rate from the lowest quintile because at an SNR of 4 the models often produced zero false negatives for smaller percentiles, making it impossible to see changes over time or to effectively compare them.

\begin{figure}[b]
	\begin{center}
		\includegraphics{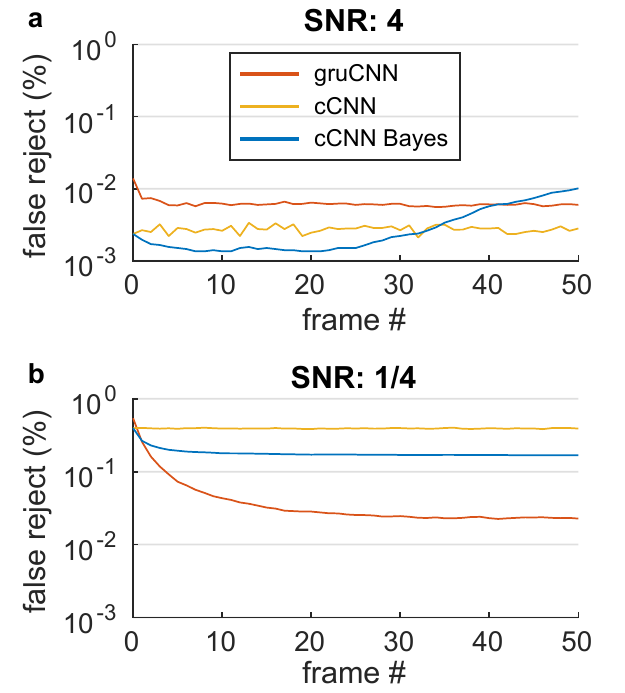}
	\end{center}
	\caption{ \textbf{False rejection rates for gruCNNs and cCNNs with and without Bayesian inference.} False rejection rate for SNRs of a) 4 and b) 1/4, in \% and logarithmic scale respectively. False rejection rates measure how often there was the positive category in the models lowest quintile (20\%) of predicted probabilities. 0\% would indicate perfect model predictions; 10\%, that the model performs at chance (vertical axes only show 0.001\% to 1\%).}
	\label{fig:quintile}
\end{figure}

Note the vertical axis scale in Figure \ref{fig:quintile}, which shows that all models performed well below chance (10\%). For the high SNR condition (Figure \ref{fig:quintile}a), the cCNNs without Bayesian inference rarely had a false rejection, performing slightly better than the gruCNNs. The cCNNs with Bayesian inference however, which had a better classification performance for every frame except the first at which they are equal (Figure \ref{fig:manySNR}a), predicted more false negatives at later frames (Figure \ref{fig:quintile}a blue line). We believe that a slight model bias against a given category in an image sequence is compounded over time and results in unwarranted confidence in the predictions that that category was not shown. This illustrates how Bayesian inference, though useful to increase the classification performance, can cause overly extreme predicted probabilities and hurt other performance measures.

For the low SNR condition, the gruCNNs substantially outperformed both cCNNs, with and without Bayesian inference (Figure \ref{fig:quintile}b). At frame number 50, there were 0.39\% and 0.17\% of false negatives in the bottom quintile of predictions from the cCNNs without and with Bayesian inference, respectively. That compares to 0.022\% of false negatives for gruCNNs, a change of one order of magnitude. These rejection rates indicate that when image quality is low, a category rejection from gruCNNs is more reliable.

\section{Performance calibration}
Next, we investigated whether the predicted probabilities are well calibrated. A well calibrated model that predicts the probability 0.4 for 100 different test sequences, should be correct approximately 40 times. Recent advances in deep neural network architectures and training, while increasing performance, have often worsened their calibration \cite{Guo2017}, highlighting the independence of these measures.

\subsection{Reliability for predicted probabilities}
\begin{figure}
	\begin{center}
		\includegraphics{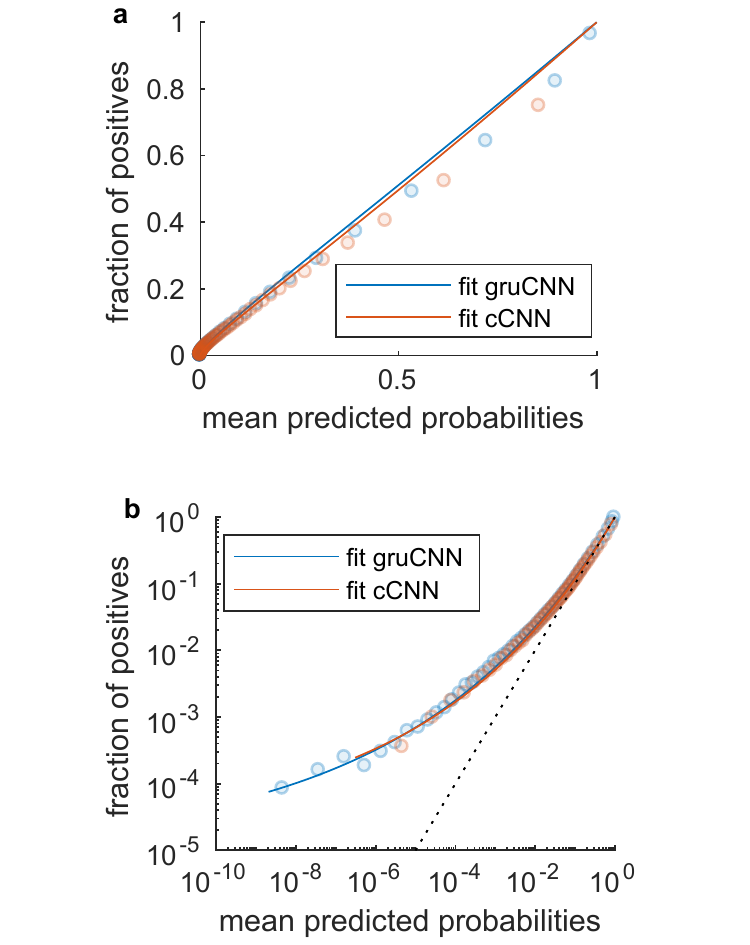}
	\end{center}
	\caption{\textbf{Reliability plots for detailed comparison of predicted probabilities (SNR = 1/4).} The plots show the same data, but the axes are scaled a) linearly and b) logarithmically. The horizontal axes show values predicted by gruCNNs and cCNNs (blue and red circles, respectively), grouped into 50 intervals of 2 percentiles each. The vertical axes display  mean occurrence rate of the image for the mean predictions value for that percentile---the fraction of positives. The red and blue lines are fits (least-squares method), used to calibrate the probability values. These fits are nearly linear at values close to one and asymptotically approach a low fraction of positives baseline for smaller predicted probabilities.}
	\label{fig:calibration}
\end{figure}
 We chose the SNR condition of 1/4 at which the gruCNNs have higher classification performance compared to cCNNs and examined how the models’ calibration metrics differed, if at all. Given infinite data and training, the output of models trained with cross-entropy should resemble the Bayesian \textit{a posteriori} probabilities \cite{Richard1991,Kline2005}, resulting in a linear relationship between the mean predicted probabilities for probability intervals and the fraction of positives for each interval. We determine the models’ calibration using reliability plots \cite{DeGroot1983,Niculescu-Mizil2005}; these show the calculated fraction of positives for a range of predicted probability intervals. We analyzed predicted probability values in two percentile increments, resulting in 50 intervals (Figure \ref{fig:calibration}). We found that for mean predicted probabilities values above 0.1, both models’ generally matched the fraction of positive image category occurrences (Figure \ref{fig:calibration}a). For low mean predicted probabilities ($\ll0.1$) however, both models’ were overly confident, the deviation from a linear relationship is clearly visible when plotted on a logarithmic scale (Figure \ref{fig:calibration}b). It appears that the log fraction of positives $\log{y}$ has a minimum for very small log probabilities  $\log{p}$, we therefore decided to fit the following exponential calibration function (least-squares fit) to the log probabilities $\log{p}$:
\begin{equation}\label{eq:cal}
\log{y}=a(1-e^{c\log{p}})
\end{equation}
 $\log{p}$ is the log fraction of positives, $a$ the minimum log fraction of positives and $c$ determines the slope. The function fit the data extremely well---explained variance is $r^2=0.998$ and $r^2=0.997$ for gruCNNs and cCNNs, respectively.
 The values for the probability minima were not statistically different, $a=-11.6=log(9.6\times10^{-6})$ and $-10.9=log(1.8\times10^{-5})$ for the gruCNNs and cCNNs respectively. We would expect such a minima if there are images in the CIFAR-10 test set that are consistently misclassified, irrespective of noise or model. In the future we plan to use larger databases with more training and test images to test whether equation \eqref{eq:cal} can be used in general to calibrate output of CNNs. Here, we can use the calibration function---and normalization to ensure probabilities sum to one---to transform the models’ predicted probability values to calibrated probabilities and analyze these further.

 We want to emphasize the importance of investigating calibration using log probabilities. Often, the calibration error is calculated by averaging the mismatch between predicted probabilities and fraction of positives on a linear scale \cite{Guo2017,Neumann2018}. The latter case results in the calibration error stemming mainly from miscalibration at high probabilities. The mean for the lowest two percentiles of predictions from the gruCNNs is $4\times10^{-9}$, the corresponding fraction of positives is $8\times10^{-5}$. In critical applications such as pedestrian detection \cite{Neumann2018} a miscalibration of a factor larger than 10,000 is highly significant, yet would not meaningfully contribute to the calibration error.

\subsection{Cumulative distribution functions for \mbox{calibrated} probabilities}
\begin{figure}
	\begin{center}
		\includegraphics{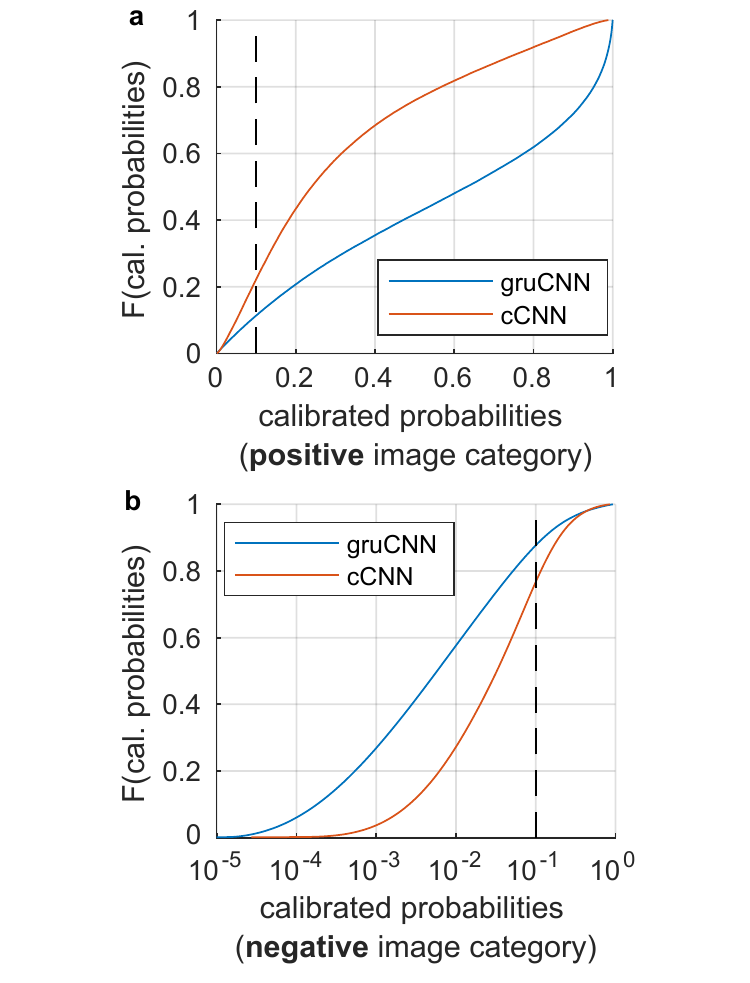}
	\end{center}
	\caption{\textbf{Cumulative distribution function (cdf).} a) cdf for the calibrated probabilities for all correct image categories (frame 50). Extremely high performing models would only predict values close to one. b) cdf for calibrated probabilities of wrong image categories. Ideal models would always predict extremely small values. Note, horizontal axis is scaled logarithmically. Blue and red lines represent probabilities for gruCNNs and cCNNs, respectively. Dashed lines indicates performance at chance.}
	\label{fig:cdf}
\end{figure}
Next, we looked at the cumulative distribution functions (cdf) over the calibrated probabilities sub-selected for the correct (positive) image category (Figure \ref{fig:cdf}a) or wrong image category (Figure \ref{fig:cdf}b). Using cdfs to plot the output of the calibration function against the ratio of values that is smaller than a given probability demonstrates the models’ performance, \ie a model with excellent performance should predict extremely high or low values for positive and negative categories, respectively. For the positive image class, 64.5\% of calibrated probabilities from the gruCNNs are above 0.4, which is twice as many compared to the cCNNs (31.5\%). Furthermore, as seen in the cdf for negative image categories (Figure \ref{fig:cdf}b): 57.6\% and 27.2\% of calibrated probabilities are below 0.01 for the gruCNNs and cCNNs, respectively. These data reveal the gruCNNs’ higher confidence in both detecting the correct image class and firmly rejecting wrong image classes at low SNRs.

\section{Alternative model architectures}
So far, we have only compared model architectures with four convolutional layers or recurrent convolutional layers. Prior to the detailed comparison of cCNNs to gruCNNs above, we tested performance of various other architectures. Details can be found in the supplementary material \ref{SupplMat}, but a short summary of relevant results follows. First, replacing fully connected layers with recurrent layers in the last stages of the CNN does not provide the same benefits as adding recurrent convolutional layers (see Supplementary Figure \ref{fig:cCNNvsgruCNN}). Second, splitting the convolutional layers into classic convolutional and recurrent convolutional layers does not significantly improve performance (see Supplementary Figure \ref{fig:whichConvLayer}). Finally, the GRU convolutional layers used above were deemed a good choice (see Supplementary Figure \ref{fig:whichReccurency}), because their performance was matched only by LSTM convolutional layers \cite{Shi2015}. We selected GRU convolutional layers over LSTM convolutional layers because LSTM convolutional layers have more parameters. The gruCNNs outperformed vanilla recurrent convolutional networks and recurrent gated inspired convolutional networks \cite{Nayebi2018}.

\section{Conclusions}
We believe that classical feedforward CNNs, although excellent for high contrast image classification, will be insufficient to meet the demands of growing technology such as self-driving cars or any real-time imaging, object classification. In these scenarios, image quality is often poor because of low light levels, and self-motion and/or object motion do not allow for compensatory long exposure times. Like others \cite{Spoerer2017,Nayebi2018}, we took inspiration from the function of the primate visual system and modified CNNs to include recurrent processing. Our goal was to improve image classification, specifically in a low SNR regime from which significant real-time data will be derived. Recurrent processing at early convolutional stages of a deep neural network enhanced performance and confidence when faced with image sequences at low SNRs.

Adversarial attacks can “trick” CNNs into misclassifying images through minimal changes to the image \cite{Biggio2013, Szegedy2013}. Humans without a viewing duration constraint are immune to adversarial image modifications, as they are regarded as the ground truth for image classification. Notably, if humans are experimentally limited to view images for short time periods (\eg less than 75 ms)---when perception is dominated by feedforward processing---they are also susceptible to adversarial attacks \cite{Elsayed2018}, again implying the important functional role for recurrent connectivity. We speculate that recurrent connections in processing of images that slightly vary might also aid in making networks more robust against adversarial attacks, something we plan to test in the future. Other future work will address the relatively small images ($32\times32$ pixels) we tested. It will be important to scale up the input and use images from databases such as ImageNet or noisy video sequences.
\section*{Acknowledgment}
We thank Anna Hartmann, Matthias Minderer, and Jan Drugowitsch for comments on the manuscript. This work was supported by NIH and an NVIDIA hardware grant.

{\small
\bibliographystyle{ieee}
\bibliography{rcnn}
}

\setcounter{equation}{0}
\setcounter{figure}{0}
\setcounter{table}{0}
\setcounter{section}{1}

%\setcounter{page}{1}
%\makeatletter

\renewcommand{\theequation}{S\arabic{equation}}
\renewcommand{\theHequation}{S\arabic{equation}}
\renewcommand{\thefigure}{S\arabic{figure}}
\renewcommand{\theHfigure}{S\arabic{figure}}
\appendix

%\xdef\presupsections{\arabic{section}}
%\renewcommand{\bibnumfmt}[1]{[S#1]}
%\renewcommand{\citenumfont}[1]{S#1}
%\SupplementaryMaterials
\pagebreak
\onecolumn
\section{Appendix}
\label{SupplMat}
\begin{figure}[!h]
	\begin{center}
		\includegraphics{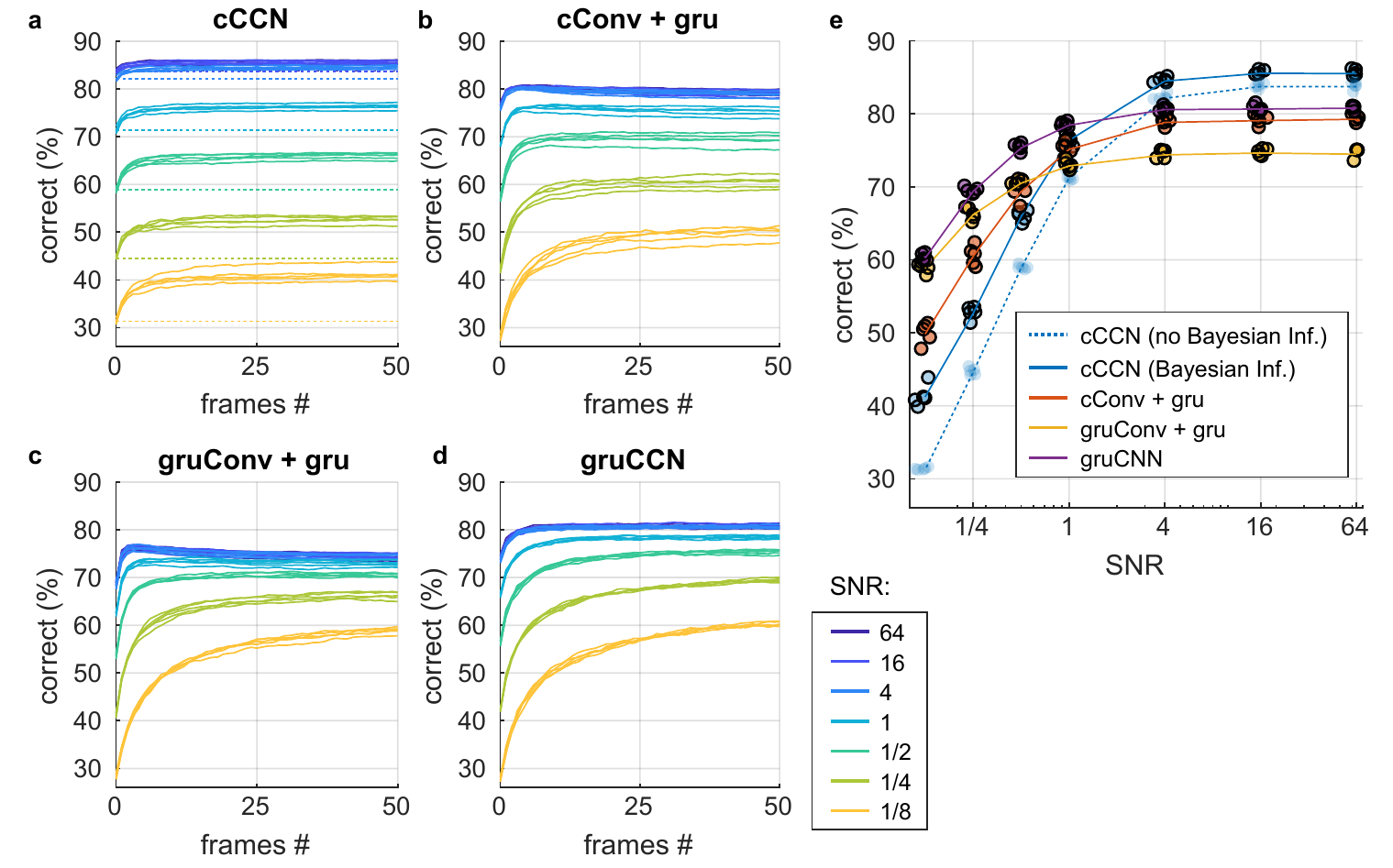}
		\caption{
			 \textbf{Difference in classification performance for four different architectures.} a-d show classification performance improves over time for image sequences tested across a range of SNRs (color coded) for all tested networks following default training paradigm. Architectures used were: a) classical feedforward CNN (cCNN) with Bayesian inference but no recurrent connection in the neural network, b) convolutional layers are unchanged but the fully connected layers replaced with GRU \cite{Cho2014} recurrent network layers, c) convolutional layers replaced by GRU convolutional layers (see equations \ref{eq:gru1}--\ref{eq:gru4}) and fully connected layers replaced by GRU,  
			 d) GRU convolutional layers but no recurrent connections in the fully connected layers. 
			 e) Classification performance for last frame (50) for network architectures shown in a-d), we also included classification performance for cCNNs without Bayesian inference (blue dashed line).
		 }
		\label{fig:cCNNvsgruCNN}
	\end{center}
\end{figure}
\subsection{Comments -- Supplementary Figure \ref{fig:cCNNvsgruCNN}}
The classification performances for cCNNs and gruCNNs were discussed in detail in the main part of the manuscript. This supplementary figure shows that replacing the fully connected layers with GRU layers \cite{Cho2014} improved classification performance at low SNRs (Figure \ref{fig:cCNNvsgruCNN}b), but significantly less than replacing the convolutional layers (Figure \ref{fig:cCNNvsgruCNN}e, red vs. purple). The neural networks in which both convolutional and fully connected layers had GRU recurrency added performed almost as well as the gruCNNs at low SNRs but worse performance at high SNR levels over time (Figure \ref{fig:cCNNvsgruCNN}c).
\\
\\
\\
\\
\\
\\
\\
\\
\\
\\
\\
\\
\\
\pagebreak
\begin{figure}[!h]
	\begin{center}
		\includegraphics{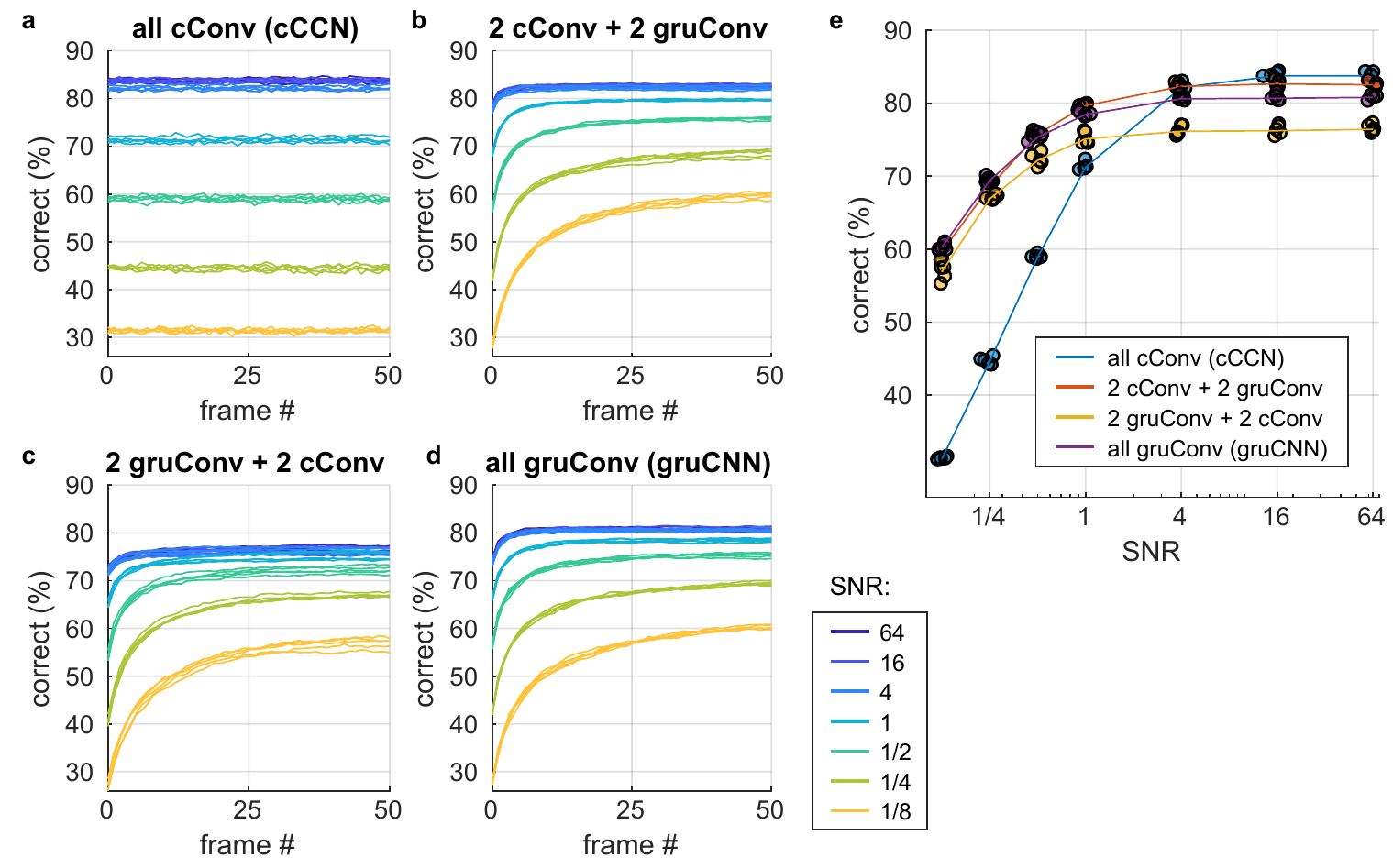}
		\caption{ 
			\textbf{Recurrent connections within the convolutional layers.} Here we show classification performance when we subselect from the four convolutional layers and replace them with GRU convolutional layers. The figure has the same format as Suppl. Figure \ref{fig:cCNNvsgruCNN}. For details of model architecture, see Table \ref{tab:model}. a) shows the classification performance for the cCNNs, but without Bayesian inference. b) The third and fourth convolutional layers were replaced with GRU convolutional layers. c) We replaced the first two convolutional layers with GRU convolutional layers.  d) Presents the same data as Suppl. figure \ref{fig:cCNNvsgruCNN}d, classification performance for the gruCNNs, in which all four are GRU convolutional layers.
		}
		\label{fig:whichConvLayer}
	\end{center}
\end{figure}
\subsection{Comments -- Supplementary Figure \ref{fig:whichConvLayer}}
The differences at low SNRs among the CNNs with recurrent convolutional layers were relatively small. The gruCNNs, in which all convolutional layers have recurrent connections, had the highest classification performance at an SNR of 1/4 and below, however the neural networks with a mix of layers---in either arrangement---were only slightly worse at low SNRs (Figure \ref{fig:whichConvLayer}e red and yellow lines vs. purple line). Additionally, the arrangement of two feedforward layers followed by two recurrent convolutional layers outperforms the gruCNNs at high SNRs (Figure \ref{fig:whichConvLayer}e red vs. purple line). In the future, we would like to identify the best arrangement of feedforward and recurrent convolutional layers with an extensive hyperparameter search. 
\\
\\
\\
\\
\\
\\
\\
\\
\\
\\
\\
\\
\\
\\
\pagebreak
\begin{figure}[!h]
	\begin{center}
		\includegraphics{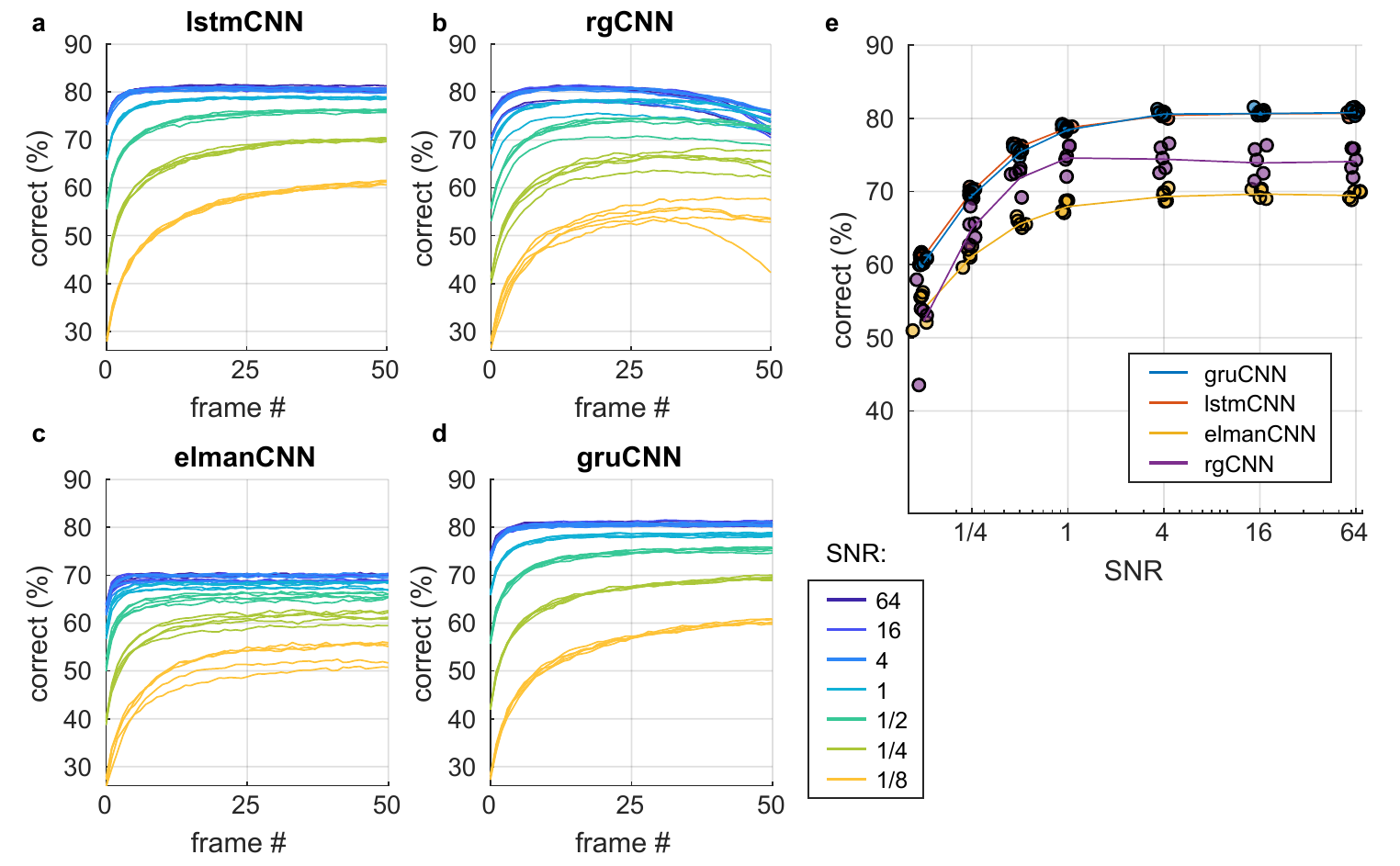}
		\caption{ \textbf{Image classification performance differences between various recurrent convolutional layer types.} The figure has the same format as Suppl. Figure \ref{fig:cCNNvsgruCNN} and \ref{fig:whichConvLayer}. a--d) Classification performance over time for four different recurrent convolutional layer types.  All architectures have the same number of channels, but different numbers of parameters. d) Plot repeated for easier comparison, same as Suppl. Figures \ref{fig:cCNNvsgruCNN}d and \ref{fig:whichConvLayer}d.}
		\label{fig:whichReccurency}
	\end{center}
\end{figure}
\subsection{Comments -- Supplementary Figure \ref{fig:whichReccurency}}
All CNNs in Suppl. Figure  \ref{fig:whichReccurency} have the same overall architecture, but with different recurrent connections within a convolutional layer. The \textbf{lstmCNN} was developed by  Shi \etal\cite{Shi2015}, we used their implementation. The \textbf{rgCNN} is an adaptation of Nayebi \etal \cite{Nayebi2018} recurrent gated convolutional networks, we used the following equations to calculate the hidden states $h_t$ and $c_t$:
\begin{equation}\label{eq:rg1}
h_t =  \Big(1-\sigma \big(W_{ch}*c_{t-1}\big)\Big) \circ \big(W_{xh}*x\big) + 
\Big(1-\sigma \big(W_{hh}*h_{t-1}\big)\Big) \circ \big(W_{h}*h_{t-1}\big) 
\end{equation}
\begin{equation}\label{eq:rg2}
c_t =  \Big(1-\sigma \big(W_{hc}*h_{t-1}\big)\Big) \circ \big(W_{xc}*x\big) + 
\Big(1-\sigma \big(W_{cc}*c_{t-1}\big)\Big) \circ \big(W_{c}*c_{t-1}\big) 
\end{equation}

For the \textbf{elmanCNN} we modified a vanilla recurrent neural network (Elman 1990 \cite{Elman1990}) to apply recurrency only locally via convolutions. The hidden state  $h_t$ is the convolved input plus the convolved previous state: 
\begin{equation}\label{eq:elman}
h_t =  \Big(\sigma\big( W_h*h_{t-1}\big) + W_{x}*x  \Big) 
\end{equation}

We studied the gruCNNs' performance in detail, because: 1) rgCNNs and elmanCNNs did not perform as well as the gruCNNs, and 2) while lstmCNNs had similar performance, they have more parameters which leads to slower training and testing. These data show, that independent of recurrent architecture, recurrent convolutional layers can dramatically improve classification performance at low SNRs compared to feedforward cCCNs.

\end{document}